# Handle Anywhere: A Mobile Robot Arm for Providing Bodily Support to Elderly Persons

Roberto Bolli, Jr., Paolo Bonato, and Harry Asada, *Fellow, IEEE*

*Abstract*— Age-related loss of mobility and increased risk of falling remain important obstacles toward facilitating aging-in-place. Many elderly people lack the coordination and strength necessary to perform common movements around their home, such as getting out of bed or stepping into a bathtub. The traditional solution has been to install grab bars on various surfaces; however, these are often not placed in optimal locations due to feasibility constraints in room layout. In this paper, we present a mobile robot that provides an older adult with a handle anywhere in space - "handle anywhere". The robot consists of an omnidirectional mobile base attached to a repositionable handle. We analyze the postural changes in four activities of daily living and determine, in each, the body pose that requires the maximal muscle effort. Using a simple model of the human body, we develop a methodology to optimally place the handle to provide the maximum support for the elderly person at the point of most effort. Our model is validated with experimental trials. We discuss how the robotic device could be used to enhance patient mobility and reduce the incidence of falls.

## I. INTRODUCTION

Over 750,000 adults aged 65 or older have died of COVID 19 [1]. The pandemic has severely impacted all in-person eldercare services, including assisted living facilities, visiting nurses, and home care; people have lost care services and community interactions. In addition to the death toll, many have suffered from mental disorders due to prolonged isolation [2]. The current work was motivated by the need for delivering high-quality eldercare services in a manner that is pandemic resilient.

Roughly 25 million Americans rely on help from caretakers and use assistive devices such as canes, raised toilets or shower seats to perform essential daily activities [3]. Falls represent a major risk, especially for isolated seniors, as the vast majority of falls occur when an elderly person is alone [4]. Thirty percent of people over the age of 65 fall each year, and falls are listed as a contributing factor to admissions to nursing homes in 40% of cases [5]. In a hospital study, almost 80% of patients who fell were unassisted, and 84.7% of total falls happened in the patient's room [4]. Lost balance was the prevailing reason given by patients, and the most common activities at the time of a fall were ambulation, getting out of bed, and sitting down or standing up – all activities requiring significant changes in body posture [5].

Roberto Bolli, Jr. is with the Department of Mechanical Engineering, Massachusetts Institute of Technology, Cambridge, MA 02139 USA (phone: 502-777-2640; e-mail: rbolli@mit.edu).
Paolo Bonato is with the Harvard Medical School Department of Physical Medicine and Rehabilitation at Spaulding Rehabilitation Hospital, Charlestown, MA 02129 USA (e-mail: pbonato@mgh.harvard.edu).
Harry Asada is with the Department of Mechanical Engineering, Massachusetts Institute of Technology, Cambridge, MA 02139 USA (e-mail: asada@mit.edu).

TABLE I. COMPARISON OF ASSISTIVE HOME DEVICES FOR THE ELDERLY

| Device | Uses | Limitations |
|---|---|---|
| Transfer sling | Supports transfers to/from a bed, wheelchair, car, etc. | Requires a human to operate |
| Patient lift (Hoyer lift) | Supports transfers in and out of a bed and/or bath. | Expensive; narrowly tailored for specific tasks |
| Walk-in shower | Allows an elderly person to safely enter a shower | Expensive; requires home renovation |
| Grab bar | Provides support for various activities | Placement constrained by room layout; only provides support in vicinity of bar |

Existing elderly assistive devices are effective for specific use cases, but their applications are often limited, as shown in Table 1. Some devices, such as transfer slings, require another person to set up and deploy, and are thus of limited use outside of institutional care settings. In addition, most are tailored for only a specific task or set of tasks. A patient lift - also known as a Hoyer lift - can be used by seniors to get in and out of bed without the assistance of another person [6], but offers no help with toileting, ambulating, or navigation. Barrier-free home improvement provides elderly people with various supports, encompassing everything from widening doorways to installing stair lifts and replacing bathtubs with walk-in showers [7]. While this is a good option for placing multiple assistive devices around the home, each tailored to a specific activity, the cost of such a treatment is a formidable barrier to most seniors, who have a median retirement income of $47,357/yr [8].

Perhaps the most widely used household balance and transfer aid are grab bars, which are often prescribed to seniors to compensate for age-related deficits [9]. These are handlebars installed in critical locations – such as near the bathtub, on each side of the toilet, and next to doorways – that elderly people can grab for bodily support. On average, each senior installs two grab bars, especially in the bathroom, where 87% reported using them for assistance on a regular basis [9]. Besides providing assistance with various tasks, they have also been shown to reduce the incidence of falls in certain scenarios [10]. However, the placement of grab bars is a major challenge, since they must be rigidly attached to a nearby surface and are therefore constrained by the room layout. This sometimes leads to inappropriate bar locations for diverse activities. Since the bars are fixed, they must be installed in every high-risk area, which is often costly, and once the user is finished using a grab bar for assistance, it provides no further support for other activities. A moveable tension pole has been proposed as a means to assist with both walking and standing [11], though this device requires a continuous flat ceiling and cannot travel through doorways.

The necessity for physical support, both to reduce falls and to improve quality of life, highlights the need for a comprehensive assistance system that can be deployed to help elderly persons navigate the home environment. The goal of the current work is to extend the functionality of the widely used grab bars to one that can be placed anywhere within the home. We propose to use a mobile robot with a repositionable handlebar that can provide a point of support for various activities requiring postural change, including ambulation, sit/stand transfers, and toileting. The support can be both physical (through offloading body weight onto the handle) and cognitive, as previous research has shown that providing contact cues at the fingertip can reduce postural sway by 50-60% [12]. By placing the handlebar effectively based on the user's body pose, we hope to emulate the assistance given by a human caretaker, which remains the gold standard in eldercare. Unlike a human caretaker, the robot can provide every elderly person with personalized assistance 24/7, which is especially important given that 58.5% of falls occur at night (between 7 p.m. and 7 a.m.) [4].

We must note, however, that safe and high-quality care services cannot be reliably provided with a fully autonomous system. A human must be in the loop to monitor the robot and control its movements if necessary. The objective of this paper is therefore twofold. One is to provide an older adult with a multi-purpose physical support system. Specifically, we focus on a robotic handlebar. The other objective is to make the support system pandemic resilient. We aim to reduce in-person care services while allowing a caregiver to access older adults remotely. Here, we develop a semi-autonomous robotic support system where a remote caregiver can monitor, operate, and intervene in the support services of the robot.

In the following, we will describe the design concept of a remotely controllable robotic support system, Handle Anywhere, and address where to place the handlebar for effectively supporting the older adult. An algorithm will be developed for determining an optimal location of the handle to provide the maximal support for a group of common tasks. The proposed method of handle placement is tested experimentally using the robotic system, and the efficacy is evaluated based on quantitative metrics (e.g. force exerted on the handlebar) and qualitative feedback from the user. Furthermore, the remote operation of the system is demonstrated with a professional caregiver accessing the robot from a hospital.

## II. ROBOT DESIGN AND IMPLEMENTATION

Our general idea is to extend the concept of grab bars such that they can be positioned anywhere in the home. We developed a list of functional requirements for the design and physical dimensions of the robotic system based on the characteristics of our target population, which consists of elderly people who require support and mobility assistance. These requirements reflect the interrelated goals of utility, feasibility, and technological acceptance, and can be separated into stipulations for the physical construction of the robot (first three bullet points) and its control and operation (final two bullet points).

- Provides both haptic and body-weight support via a handlebar
- Handlebar can be positioned arbitrarily, to assist with different activities of daily living
- Can navigate the standard home environment
- Human-in-the-loop: for safety, robot must be controlled by a caretaker if necessary
- Capable of remote teleoperation, so a caretaker is not required to be physically present

To further determine design specifications, we consider typical use scenarios and environment conditions, as shown in Fig. 1. The robot must be able to maneuver thorough a confined space such as a bathroom or a bedroom. If the elderly person is in bed and desires to sit up, we would like to be able to place the handlebar based on his or her current body position (Fig. 1, top left). The handle should also be able to assist the user during ambulation sequences; for example, by helping the user stand up in a bathtub (green-colored pose, Fig. 1, top right) and then step over the bathtub lip (normal-colored pose). In addition, the system should provide the caretaker with the ability to execute a sequence of steps for a complex motion, such as getting out of bed. We provide an example sequence of four unique handlebar poses for this movement in the bottom of Fig. 1, including lie-to-sit support, sit-to-stand support, and assistance ambulating around the room. The corresponding body poses of the elderly person at each point in the sequence are shown on the right.

These functional requirements were realized through our implementation of the robotic system, which consists of a 6-DOF Universal Robotics UR10e arm mounted on a custom-made omnidirectional vehicle with four Mecanum wheels (Fig. 2). The flat, U-shaped base was designed to fit underneath common bed and table configurations, and provides a space for the elderly person to stand in. At the end of the robot arm, we attached a T-shaped handlebar instrumented with a 6-DOF force/torque sensor and embedded grip sensors. The use of a UR10e allowed the handle to be placed in any arbitrary position and orientation. Both the vehicle and the handlebar were padded with thick foam to reduce the chance of injury; in the case of the handle, this also served to increase grip friction and prevent the user's hands from slipping. The dimensions of the drive base and handle, presented in Table 2, were chosen based on empirical ergonomics research and typical home layout constraints. In addition, the UR10e control box and all power equipment were mounted on a battery powered mobile cart to make the robot more compact and maneuverable.

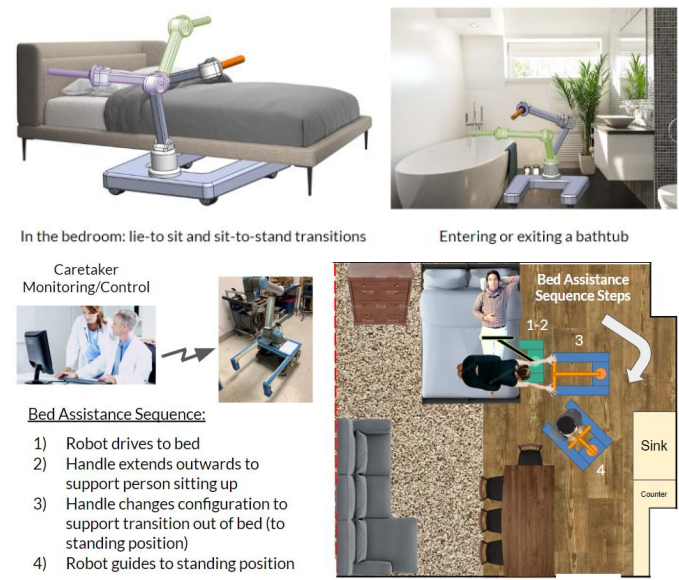

Figure 1. Robot usage scenarios (top) and example bedroom assistance sequence (bottom).

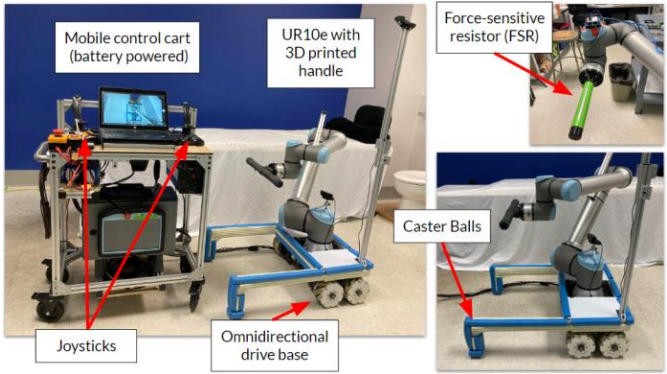

Figure 2. Handle anywhere robot system.

TABLE II. DESIGN CONSIDERATIONS FOR THE ROBOT DIMENSIONS

| Dimension | Value | Rationale |
|---|---|---|
| Drive base width | 66 cm | Most common home doorways are ≥ 71 cm (28") wide [13] |
| Drive base length | 84 cm | Allows for a 48 cm U-shaped drive base for elderly person to stand/walk in |
| Handle length | 46 cm | Sufficient for a two-handed grip |
| Handle diameter | 3.8 cm | Ergonomics studies suggest an optimal range of 3.56-4.06 cm [14] |
| Handle reach from robot | 44 cm | Limited to this value to prevent the robot from tipping |

To satisfy the requirements for versatility and human-in-the-loop control, as well as for remote teleoperation, we adopted a semiautonomous control scheme where the robot movement was overseen and controlled by a human operator with various tools (Fig. 3), in a form of human supervisory control [15]. This scheme allows a caretaker to monitor/assist a patient's movements remotely, and tune the system when they were present in person. We envision this teleoperation paradigm as a step towards pandemic-resilient eldercare, since our system enables a caregiver to physically support elderly users without having to be present. Four camera views (two from the robot, two from the cart) along with a graph of the 6-axis force and torque data, grip strength on the left and right of the handle, and net applied force (Fig. 4) were transmitted to a remote computer. The cameras were mounted to provide a front view of the patient as well as a wide-angle view of the robot and its surroundings. Joysticks were used for human control of the robot. The operator could change the handlebar's height, distance from the center of the robot arm, and rotation relative to the mobile base. The actuators were controlled by either position or velocity control, with a safety stop to prevent injury to the user. Since the mobile base allowed for holonomic movement, each degree of freedom was mapped to a 3-axis joystick (the $3^{rd}$ axis being the rotation of the joystick itself). In addition, the robot had the capability to switch to "freedrive" mode, whereby the handlebar could be manually positioned by an in-person caretaker. This allowed for the user's preferred handlebar placements to be saved and retrieved from memory.

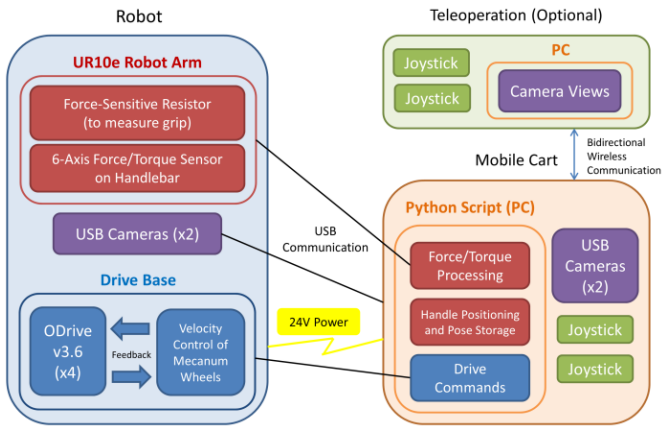

Figure 3. Block diagram of the robotic system.

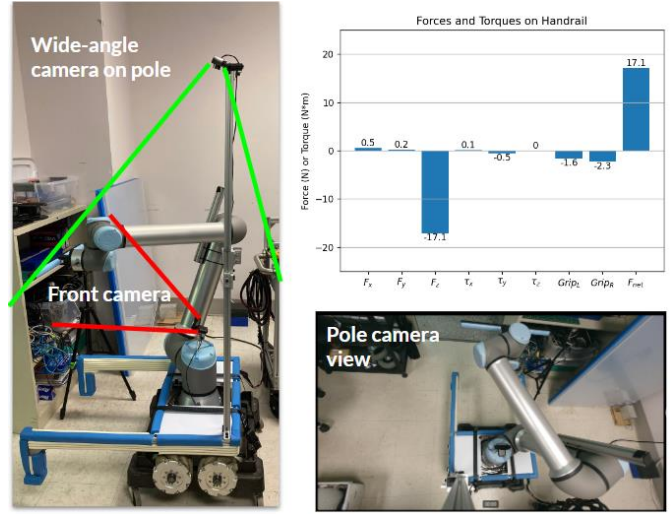

Figure 4. Caretakers' view during teleoperation.

A live bidirectional audio and visual link enabled the caretaker to communicate with the user and receive consent for each movement. Depending on the user's feedback, the caretaker could modify the handle position to better support the patient. Continuous force and grip monitoring would alert the operator when the user grabbed or released the handle, and if the net handlebar force exceeded the payload of the UR10e, the system would enter a protective stop.

While the robot hardware allowed for the handlebar to be used to apply a force on the user (e.g. pulling them up from a chair), we constrained the handlebar to be completely stationary while the user grabbed onto it. From our experience, elderly adults tend to be afraid of assistive devices that might move when they are not expecting it, as they fear such a movement might cause a fall or a slip. Since our target users are expected to shift a significant amount of their body weight onto the handlebar, we believe that a stationary handle will be perceived as safer and more trustworthy, especially since elderly adults are already familiar with grab bars. Therefore, to increase adoption of the robot system and to eliminate any possibility of the robot triggering a fall, we decided to rigidly fix the handlebar in place while it actively supported the user.

The robot system was tested remotely with a physical therapist at Spaulding Rehabilitation Hospital, and was confirmed to work as a proof-of-concept. The physical therapist was able to successfully drive the robot and position the handlebar to assist with toileting and bathing. Future work involves eliminating the mobile cart so that the robot can be truly untethered. Additionally, we aim to investigate methods to safely move the handrail while it is being grabbed, so the robot can actively move or reposition the user.

### III. OPTIMIZATION OF HANDLE LOCATION

In order to maximize the utility provided by the handle anywhere robot, we developed a mathematical model to position the handlebar based on the body pose requiring the most muscle effort for the activity the user is performing. As explained in section II, we only consider the scenario where the handle is stationary. We begin by modeling the human body as a 7-bar linkage (Fig. 5), confining body movement to the sagittal plane and assuming that both arms move simultaneously. Human arm muscle effort is represented via joint torques $\tau_5$, $\tau_6$, and $\tau_7$, acting on $\theta_5$, $\theta_6$, and the end of link 6 (at the origin of frame $x_7, y_7$), respectively. We use the joint representation described by Hatsukari et al. [16] with a slight modification: both $\theta_4$ and $\theta_5$ are measured from the coordinate frame $x_4, y_4$ so that the arm and head angles are relative to the trunk. Each of the links was given a mass based on the physical composition of the corresponding part of the human body [17], with some links absorbing the mass of multiple body parts (Fig. 5). For this study, the mass of each link was determined by estimation from an adult volunteer (23 years old, 60 kg), and all masses were normalized so that the sum of the links was the total mass of the body.

Applying the 7-bar linkage model to a person holding onto a handlebar results in a closed-loop kinematic chain, which yields highly complex equations of motion and complicated force/torque interactions. To simplify this, for each activity, we chose to analyze only the body pose requiring the most muscle effort. Four scenarios were selected as representative and diverse examples of ambulation activities that elderly people have difficulty performing [18]: lie-to-sit in a bed, sit-to-stand in a bed, standing up in a bathtub, and sit-to-stand from a toilet. Our volunteer was filmed in the sagittal plane performing each scenario without any assistance. Afterwards, he identified the body poses that required the maximal muscle exertion, which are shown in Fig. 6. The applicability of this data to elderly people is discussed in section V.

Introducing a handlebar presents another challenge for our analysis: during each motion, if the user grabs onto the handlebar, the body pose requiring the maximal muscle effort will likely be different from the poses identified in Fig. 6, as the user would change the location of his/her arms (and possibly trunk) to reach the handlebar. This means that the pose of most effort would be dependent on the location of the handlebar. To avoid this, we introduce two constraints: first, we limit all possible handlebar locations to be within reach of the arms (links 5 and 6) and assume that the user will grab the handle by moving only their arms. Second, we base our analysis of the body pose on the remaining five links, which account for all parts of the body except the arms. Since this encompasses almost 93% of the total body mass, little information is lost. We calculate the center of mass (COM) of these five links (1), where $x_i$ and $y_i$ are the locations of the center of mass of each linkage, and $M$ is the total mass of the test subject. The orientation and COM of the first five links during each of the four scenarios is shown in Fig. 6.

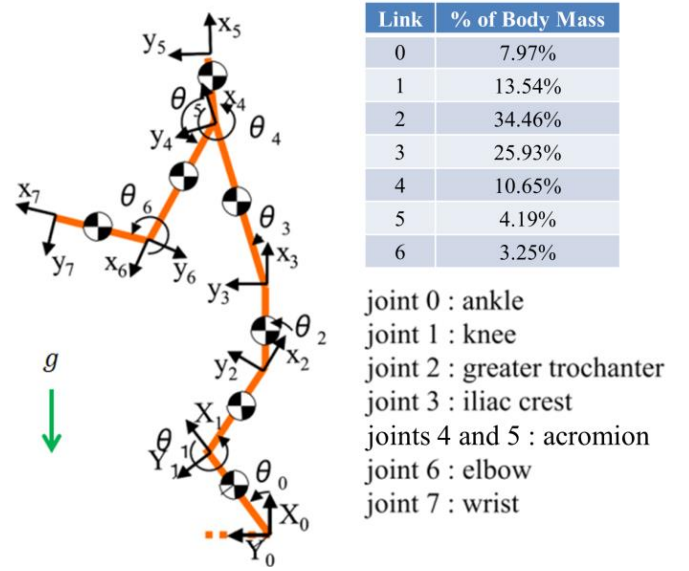

| Link | % of Body Mass |
|------|----------------|
| 0    | 7.97%          |
| 1    | 13.54%         |
| 2    | 34.46%         |
| 3    | 25.93%         |
| 4    | 10.65%         |
| 5    | 4.19%          |
| 6    | 3.25%          |

joint 0 : ankle
joint 1 : knee
joint 2 : greater trochanter
joint 3 : iliac crest
joints 4 and 5 : acromion
joint 6 : elbow
joint 7 : wrist

Figure 5. Model of human body composed of seven linkages (adapted from T. Hatsukari et al. [16]).

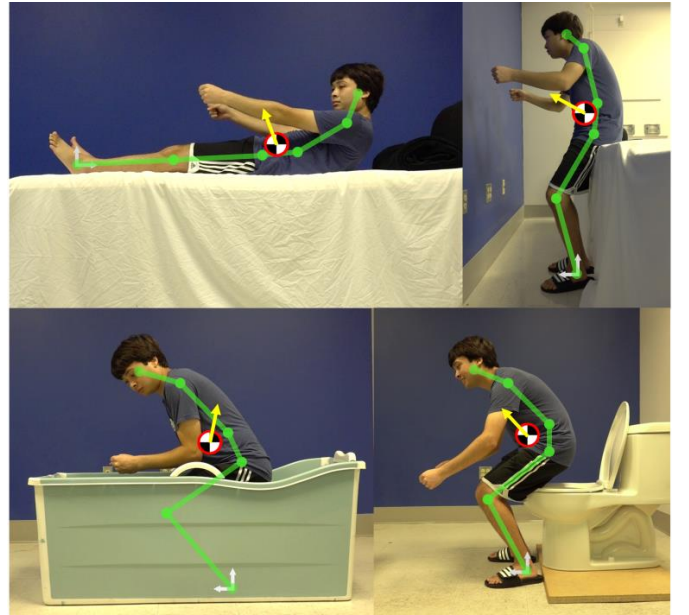

Figure 6. Human subject at points of maximal muscle exertion, with the position and normalized velocity vector of the center of mass of the first five body links (shown in green).

$$X_{COM} = \frac{1}{M} \sum_{i=0}^{4} m_i x_i \qquad (1)$$

$$Y_{COM} = \frac{1}{M} \sum_{i=0}^{4} m_i y_i$$

The normalized velocity vector of the first five links, $\overrightarrow{v_{COM}}$, was also calculated by analyzing several frames of a recording of the volunteer performing each scenario, and tracing the motion of a body marker in the frames immediately preceding and following the pose that yielded the volunteer's maximal exertion. Since any human would have some intended motion to achieve during each scenario, these two measures – COM position and velocity – give us an idea of the volunteer's instantaneous body trajectory.

As stated earlier, when the user grabs the handlebar, a closed-loop kinematic chain is formed with the handlebar and the ground. To avoid the complexity this causes, we condense the first five links in the body into a single point mass located at $(X_{COM}, Y_{COM})$ with velocity $\vec{v_{COM}}$ (Fig. 7). We can then simplify the human body model into a three-bar serial linkage consisting of links 6, 5, and a virtual link $\vec{r_{COM}}$ from link 5 to $(X_{COM}, Y_{COM})$. The origin of the linkage is located at the handlebar. Thus, given a body pose, we can calculate an acceptable handlebar location by choosing $\theta_5$ and $\theta_6$ and using forward kinematics.

With the handlebar fixed in place, an older adult can generate a force $F_{arm}$ at their COM via arm joint torques $\tau_5, \tau_6,$ and $\tau_7$ (2), allowing them to move their body in an intended direction. This force is given by the Jacobian $J$ of the three-bar linkage in Fig. 7.

$$F_{arm} = (JJ^T)^{-1} J \begin{pmatrix} \tau_7 \\ \tau_6 \\ \tau_5 \end{pmatrix} \quad (2)$$

The expanded form of (2) is shown in equation (3). For convenience in our calculations, we create a coordinate frame $x_{0,4}, y_{0,4}$ at the same origin as $x_4, y_4$ but whose axes align with $x_0, y_0$, so that $\theta_{0,4}$ is the angle between the two frames (Fig. 7). The angle $\theta_{COM}$ of $\vec{r_{COM}}$ is then defined with respect to $x_{0,4}, y_{0,4}$. Links 5 and 6 are denoted by the vectors $\vec{r_5}$ and $\vec{r_6}$.

$$F_{arm} = \quad (3)$$
$$\frac{\tau_5}{|\vec{r_{COM}}|} \begin{bmatrix} -\sin\theta_{COM} \\ \cos\theta_{COM} \end{bmatrix} + \frac{\tau_6}{|\vec{r_{COM}} - \vec{r_5}|} \begin{bmatrix} -\sin\theta_{6,COM} \\ \cos\theta_{6,COM} \end{bmatrix}$$
$$+ \frac{\tau_7}{|\vec{r_{COM}} - \vec{r_5} - \vec{r_6}|} \begin{bmatrix} -\sin\theta_{7,COM} \\ \cos\theta_{7,COM} \end{bmatrix}$$

where

$$\theta_{6,COM} = \theta_{0,4} + \theta_5 - \pi - \cos^{-1}\left(\frac{|\vec{r_5}|^2 + |\vec{r_{COM}} - \vec{r_5}|^2 - |\vec{r_{COM}}|^2}{2|\vec{r_5}||\vec{r_{COM}} - \vec{r_5}|}\right)$$

$$\theta_{7,COM} = \theta_{0,4} + \theta_5 + \theta_6 - \pi$$
$$+ \cos^{-1}\left(\frac{|\vec{r_6}|^2 + |\vec{r_{COM}} - \vec{r_5} - \vec{r_6}|^2 - |\vec{r_{COM}} - \vec{r_5}|^2}{2|\vec{r_6}||\vec{r_{COM}} - \vec{r_5} - \vec{r_6}|}\right)$$

At the desired handrail location, $|F_{arm}|$ should be as large as possible to provide the elderly person with the most assistance in moving their body, and the vectors $F_{arm}$ and $\vec{v_{COM}}$ should be aligned so that the body can be moved in the intended direction. In other words, our goal is to maximize the inner product of $F_{arm}$ and $\vec{v_{COM}}$. We also want to locate the handrail a comfortable distance away from the user's shoulder – not too close and not too far – so that the handlebar can support the user throughout the motion. Therefore, we introduce a penalty for extremely proximal or distal handrail placements, which simplifies kinematically to a relationship with $|\cos\theta_6|$. This optimization problem is formulated in (4), with the constant $a = 0.2$ chosen to weigh the distance penalty appropriately.

$$argmax(F_{arm} \cdot \vec{v_{COM}} - a|\cos\theta_6|) \quad (4)$$

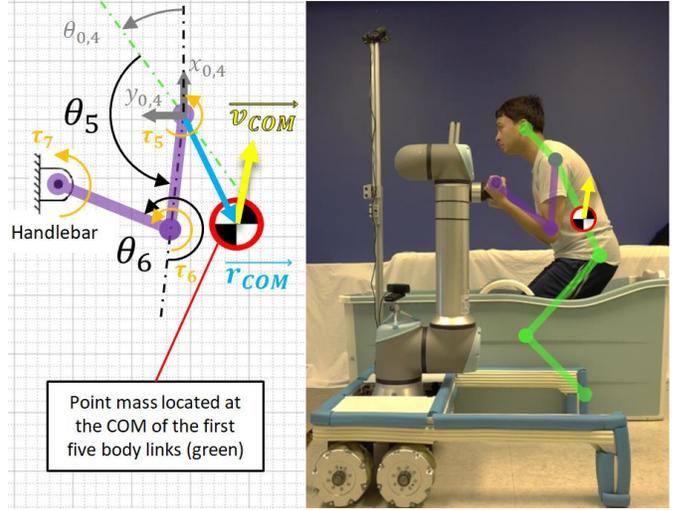

Figure 7. Diagram of the arm serial linkage (left), including the joint torques and coordinate frame $x_{0,4}, y_{0,4}$. The first five links of the body (green, right) are consolidated to a point mass located at the center of mass.

As a simplifying approximation, we assumed that the joint torques were independent of the arm angles $\theta_5$ and $\theta_6$; that is, that a human could produce a given joint torque with roughly equivalent muscle exertion when their arm is in different positions. This meant that we could treat the joint torques as constants that were either positive or negative, since in the optimization, the magnitude of the joint torque has no effect on the optimal handle position. This may sound surprising, but is due to the use of the inner product in (4). For each joint, as long as the corresponding direction of the force vector on the COM is within $\pm 90°$ of $\vec{v_{COM}}$, the inner product will be positive; therefore, in any numerical simulation to find the global maximum of (4), the torques will always take on the maximum positive or minimum negative value within the provided range. As an alternative explanation of why this occurs, we note that our assumption that joint torques are independent of arm angles means that in our model, each arm joint can produce some maximum torque independent of the arm configuration. These torques collectively act to produce a force $F_{arm}$ at the COM (the "endpoint" of the arm linkage). The effect of the inner product of $F_{arm}$ and $\vec{v_{COM}}$ is to position the arm to maximize $F_{arm}$ in the direction of the COM velocity.

Eq. 4 can be viewed as a tradeoff between the mechanical advantage and proximity of the arm configuration, which is determined by the handrail location. The mechanical advantage of the arm is defined as (5), where $|\tau|$ is the norm of the arm joint torques $\tau_5, \tau_6,$ and $\tau_7$. However, since the magnitude of the joint torques does not affect the result of the optimization in (4), as explained previously, we can pick $\tau_5, \tau_6,$ and $\tau_7$ to be constants that are either positive or negative, meaning that $|\tau|$ is also constant.

$$\frac{|F_{arm}|}{|\tau|} = c|F_{arm}| \quad (5)$$

We note that since $F_{arm}$ and $\vec{v_{COM}}$ are both real, the inner product can be written as (6), where $\theta_{VF}$ is the angle between the two vectors.

$$F_{arm} \cdot \vec{v_{COM}} = |F_{arm}||\vec{v_{COM}}|\cos\theta_{VF} = c|F_{arm}|\cos\theta_{VF} \quad (6)$$

The magnitude of $\vec{v_{COM}}$ is independent of any of the parameters we are optimizing, and remains constant for each scenario. Comparing (5) and (6), we see that through the inner product, we are essentially calculating the mechanical advantage in the direction of $\vec{v_{COM}}$.

A numerical simulation was set up in Matlab to determine the optimal joint angles $\theta_5$ and $\theta_6$ based on the cost function (4). We placed limits on the joints' range of motion to avoid solutions that were not physically achievable. To find the global optimum handlebar location, we iterated over all permissible values of $\theta_5$ and $\theta_6$ with a sufficiently small step size. For all of the scenarios, the simulation successfully converged to a unique solution; these were evaluated in section IV.

IV. EXPERIMENTAL RESULTS

Using the methodology in the previous section, we calculated a pose for the handle in each of the four scenarios. At first glance, the predicted poses appeared to be physically reasonable, directing the arm strength in the proper direction. This intuition was confirmed when the scenarios were reenacted with the handle in place, as shown in Fig. 8. In every case, the subject was able to grab the handlebar and complete the body motion naturally, following the same trajectory as without the handlebar.

Previous studies by the US military have quantified arm strength in various directions based on the degree of elbow flexion [19]. The data show that push/pull force increases as the arm extends, while up/down force peaks when the arm is bent at 90°. Looking at the COM velocity vectors in Fig. 8, we notice that the top two scenarios mainly involve rotating the body around the contact point (the buttocks), so the push/pull force should be maximized. Meanwhile, in the bottom two scenarios, the arm assists with lifting the body upwards. Thus, the handle placements yield arm configurations which optimize strength in the desired directions, in line with the military studies' measured arm strength data.

To evaluate the effectiveness of the handlebar, we took measurements of the force applied on the bar during each motion (Fig. 9). Compared to the baseline of no handle assistance, these measurements give us an idea of how much body weight the subject offloaded onto the handle. For three of the four scenarios, a significant amount of downward force was applied to the handlebar, with the arms supporting a maximum of 20-30% of the total body weight. Since the legs are able to exert roughly 4x more force than the arms [20], this means that the maximal muscle effort was relatively equally distributed between the arms and the legs. In the fourth scenario (bed lie-to-sit), the movement mainly involved rotation around the hips, so the applied force was predominantly in the horizontal direction. The horizontal forces in the other scenarios represent arm assistance towards maintaining the COM trajectory, such as pulling the arms forward to stand up.

Additionally, for the toilet scenario, we compared our calculated handlebar placement (in front of the user) to the government standard for toilet handrails (on the side). Fig. 10 shows that the test subject applied significantly more force on the side-facing bar as compared to the front-facing bar, with the arms supporting a maximum of 41% of the total body weight. This indicates that the standard toilet grab bar placement leads to a highly unequally distributed muscle effort. By contrast, our calculated front-facing handle position led to a maximal arm support of 25% of the body weight, enabling the user to leverage their leg muscles more effectively for the sit-to-stand movement.

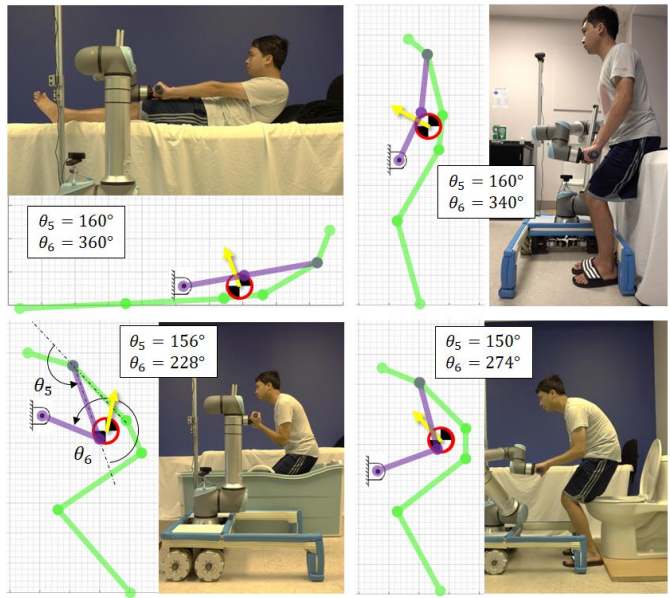

Figure 8. Image of each scenario with handle anywhere robot, including diagrams of arm angles and center-of-mass (COM) velocity.

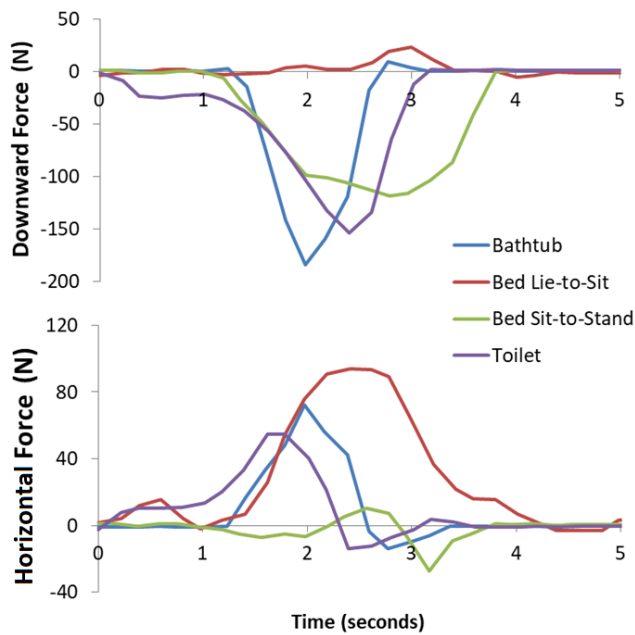

Figure 9. Downward and horizontal (antero-posterior) forces on the handlebar during each scenario.

Lastly, we asked the test subject to self-report the difficulty of executing each scenario with and without the handlebar (Table 3). This helped to reveal any qualitative differences in muscle exertion or overall patient comfort that were not captured in the force data. In all circumstances, the handlebar lowered the perceived difficulty of performing each task, moving three out of four scenarios to the "easiest" rating. The difference was most extreme for the lie-to-sit task, likely due to the intense abdominal muscle strain necessary to move the trunk to the upright position. Overall, the responses indicate that our system is likely to be readily adopted, especially by users who have difficulty performing some or all of the four scenarios we studied.

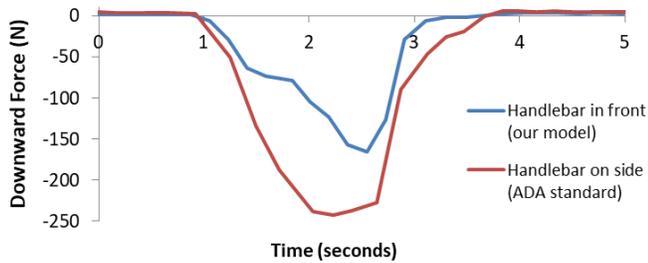

Figure 10. Downward force on the handlebar while standing up from a toilet, when the handle was located on the side (government standard) or in front of the user (our calculated handle placement)

TABLE III. SUBJECTIVE DIFFICULTY OF PERFORMING EVERYDAY TASKS. NOTE THAT 1 = EASIEST AND 5 = HARDEST.

| Scenario | Difficulty without handlebar | Difficulty with handlebar |
|---|---|---|
| Lie-to-sit in a bed | 5 | 2 |
| Sit-to-stand in a bed | 2 | 1 |
| Standing up in a bathtub | 3 | 1 |
| Sit-to-stand from a toilet | 2 | 1 |

## V. CONCLUSION

We successfully developed a mobile robot ("handle anywhere") capable of satisfying the functional requirements we identified; namely, to provide a repositionable handlebar for an elderly user and to facilitate remote monitoring and assistance. To maximize the utility provided by the handlebar, we developed a methodology to locate the bar based on the body pose requiring the highest muscle effort for the activity the user was performing. In experimental trials of four activities of daily living, the calculated handlebar locations were successful at offloading a significant portion of body weight and reducing the perceived effort required to perform each activity. We hope to employ our robot to provide bodily support to the elderly, with the goal of assisting activities requiring postural changes and reducing the incidence of falls.

The current experimental results are for a healthy young adult. It is likely that the poses of maximal effort for each activity would be different in an elderly person. However, we believe that this does not impact the validity of our methodology, as it could still be used to generate handlebar locations based on the body pose of the elderly person. To address this, we hope to conduct studies with older adults having various age-related disorders. Another limitation to the current work is our representation of muscle effort as pose-independent joint torques. A musculoskeletal model will be required to better understand the effect of the handlebar upon individual muscles [21].

We envision our technology as a step towards pandemic-resilient eldercare devices: assistive tools caregivers can use to maintain a high level of care during periods of physical isolation. Specifically, we believe that mobile handle robots can find utility at residences and nursing facilities by providing an anchor of support during postural transitions and assisting with activities of daily living.

ACKNOWLEDGMENTS

This material is based upon work supported by the National Science Foundation (NSF) Graduate Research Fellowship under Grant No. 2141064, and National Robotics Initiative Grant No. 2133075. The experimental protocol was reviewed and approved by the MIT Committee on the Use of Humans as Experimental Subjects. We thank the study participant for consenting to the unrestricted release of all photography and video recordings obtained during the study.

REFERENCES

[1] Johns Hopkins University & Medicine. "Coronavirus Resource Center 2020." https://coronavirus.jhu.edu/ (accessed September 15, 2022).

[2] World Health Organization. "COVID-19 pandemic triggers 25% increase in prevalence of anxiety and depression worldwide." https://www.who.int/news/item/ 02-03-2022-covid-19-pandemic-triggers-25-increase-in-prevalence-of-anxiety-and-depression-worldwide (accessed September 15, 2022).

[3] Amber Willink et al., "Are Older Americans Getting the Long-Term Services and Supports They Need?" *Commonwealth Fund* (2019). https://doi.org/10.26099/tdet-jr02

[4] Hitcho, Eileen B et al. "Characteristics and circumstances of falls in a hospital setting: a prospective analysis." *Journal of general internal medicine* vol. 19,7 (2004): 732-9. doi:10.1111/j.1525-1497.2004.30387.x

[5] Tinetti, M E et al. "Risk factors for falls among elderly persons living in the community." *The New England journal of medicine* vol. 319,26 (1988): 1701-7, doi:10.1056/NEJM198812293192604

[6] AssistedLiving.org. "The Best Hoyer Lifts: What Is a Hoyer Lift and How Do You Use It?" https://www.assistedliving.org/best-hoyer-lifts/ (accessed July 8, 2022).

[7] Eldercare Locator, Administration of Community Living. "Home Improvement Assistance." https://eldercare.acl.gov/public/resources/factsheets/home_modifications.aspx (accessed August 24, 2022)

[8] United States Census Bureau. (2019). "Age of Householder-Households, by Total Money Income, Type of Household, Race and Hispanic Origin of Householder." https://www.census.gov/data/tables/time-series/demo/income-poverty/cps-hinc/hinc-02.html (accessed Aug. 5, 2022)

[9] Guitard, Paulette et al. "Use of different bath grab bar configurations following a balance perturbation." *Assistive technology: the official journal of RESNA* vol. 23,4 (2011): 205-15; quiz 216-7. doi:10.1080/10400435.2011.614674

[10] Levine, Iris C et al. "Grab Bar Use Influences Fall Hazard During Bathtub Exit." Human factors, 187208211059860. 28 Dec. 2021, doi:10.1177/00187208211059860

[11] R. Hanaoka and T. Murakami, "A novel assist device for Tension Pole based movable handrail," *IECON 2015 - 41st Annual Conference of the IEEE Industrial Electronics Society*, 2015, pp. 2235-2240, doi: 10.1109/IECON.2015.7392434.

[12] Jeka, J J. "Light touch contact as a balance aid." *Physical therapy* vol. 77,5 (1997): 476-87. doi: 10.1093/ptj/77.5.476

[13] Window World of Southern Nevada. "Standard Door Width & Length Guide." https://www.windowworldsouthernnevada.com/article/standard-door-width-length-guide (accessed Aug. 5, 2022)

[14] Kose, S., Sugimoto, Y., Goto, Y. "Acceptable Handrail Diameter for Use by the Elderly." *Advances in Industrial Design. AHFE 2020. Advances in Intelligent Systems and Computing,* vol 1202. Springer, Cham. https://doi.org/10.1007/978-3-030-51194-4_28

[15] T. Sheridan, *Telerobotics, Automation, and Human Supervisory Control.* Cambridge, MA, USA: MIT Press, 1992.

[16] T. Hatsukari *et al*., "Self-help standing-up method based on quasi-static motion," *2008 IEEE International Conference on Robotics and Biomimetics*, 2009, pp. 342-347, doi: 10.1109/ROBIO.2009.4913027.

[17] Exercise Prescription on Internet. "Body Segment Data." https://exrx.net/Kinesiology/Segments (accessed Aug. 1, 2022)

[18] Edemekong PF, Bomgaars DL, Sukumaran S, et al., *Activities of Daily Living.* Treasure Island, FL, USA: StatPearls Publishing, 2022.

[19] Notices 1 and 2 Human Engineering Design Criteria for Military Systems, Equipment and Facilities, DOD, C Revision 05/02/81
[20] Climb Strong. "Leg Strength as a Limiting Factor, Revisited." https://www.climbstrong.com/education-center/leg-strength-limiting-factor-revisited/ (accessed August 24, 2022)
[21] Daniel, P., Fu, C.L., and Asada, H., "Musculoskeletal Load Analysis for the Design and Control of a Wearable Robot Bracing the Human Body While Crawling on a Floor", *IEEE Access*, 2022, vol. 10, pp. 6814-6829.